\documentclass[runningheads]{llncs}

\usepackage{eccvabbrv}

\usepackage{graphicx}
\usepackage{booktabs}
\usepackage{wrapfig}
\usepackage{hyperref}
\usepackage{url}
\usepackage{tabularx}
\usepackage{booktabs}
\usepackage{multirow}
\usepackage{multicol}
\usepackage{graphicx}
\usepackage{enumitem}
\usepackage{wrapfig}
\usepackage{caption}
\usepackage{amsmath}
\usepackage{bm}
\usepackage{amssymb}

\usepackage{orcidlink}

\begin{document}

\title{In-context Prompt Learning for Test-time Vision Recognition with Frozen Vision-language Model} 

\titlerunning{In-context Prompt Learning for Test-time Vision Recognition}

\author{Junhui Yin\inst{1} \and
Xinyu Zhang\inst{2} \and
Lin Wu\inst{3} \and
Xiaojie Wang\inst{1}}

\authorrunning{J. Yin et al.}

\institute{Beijing University of Posts and Telecommunications, China \and
Baidu Inc., China \and
Swansea University, United Kingdom}

\maketitle

\begin{abstract}

Current pre-trained vision-language models, such as CLIP, have demonstrated remarkable zero-shot generalization capabilities across various downstream tasks. However, their performance significantly degrades when test inputs exhibit different distributions. In this paper, we explore the concept of test-time prompt tuning (TTPT), which facilitates the adaptation of the CLIP model to novel downstream tasks through a one-step unsupervised optimization that involves only test samples. Inspired by in-context learning in natural language processing (NLP), we propose \textbf{In}-\textbf{C}ontext \textbf{P}rompt \textbf{L}earning (InCPL) for test-time visual recognition tasks, 
which empowers a pre-trained vision-language model with labeled examples as context information on downstream task. Specifically, InCPL associates a new test sample with very few labeled examples (sometimes just one) as context information, enabling reliable label estimation for the test sample and facilitating model adaptation. To achieve this, InCPL employs an efficient language-to-vision translator to explore the textual prior information for visual prompt learning.
Further, we introduce a context-aware unsupervised loss to optimize visual prompts tailored to test samples. Finally, we design a cyclic learning strategy for visual 
and textual prompts to ensure mutual synergy across different modalities. This enables a pre-trained, frozen CLIP model to adapt to any task using its learned adaptive prompt. Our method demonstrates superior performance and achieves state-of-the-art results across various downstream datasets.

\end{abstract}

\section{Introduction}

Recent advances in vision-language pre-training, such as CLIP~\cite{radford2021learning}, have shown a promising direction for developing foundation models for downstream tasks~\cite{bommasani2021opportunities}. These foundation models are trained on extensive web-scale data, such as 400 million text-image pairs in the case of CLIP, to align language and vision modalities. 
When transferred to downstream tasks,
the pre-trained model needs to be fine-tuned on a few labeled images for target domain.
Fully fine-tuning a large pre-trained model~\cite{bommasani2021opportunities,liu2021unified,cai2022x} for each downstream task faces two major challenges in real-world applications. One practical concern is the storage and distribution issue, as maintaining a separate model copy for each task is costly and inflexible, particularly with an expanding number of downstream tasks. Another one is that fully fine-tuning has a large probability of destroying the initial knowledge provided by the large-scale pre-trained model, increasing the risk of model overfitting.

Unlike fully fine-tuning, which expensively updates all model parameters for each downstream task, prompt tuning methods~\cite{zhou2022learning,zhou2022conditional} prepend the inputs with learnable parameters that steer the foundation model towards generating the desired outputs. Recent works~\cite{zhou2022learning,zhou2022conditional,bahng2022exploring,jia2022visual} use training data to tune prompts where the embeddings of prompt are extracted from the model input and are differentiable with respect to the loss function. However, the learned prompts are largely limited to training data, and their performance significantly degrades when the test samples are drawn from a different distribution. Test-time prompt tuning (TTPT)~\cite{shu2022test,feng2023diverse} offers a viable solution to the distribution shift problem through one-step optimization on an unsupervised objective that involves only unlabeled test samples. For example, TPT~\cite{shu2022test} exploits the pre-trained vision-language foundation models as better zero-shot learners by maximizing the consistency of predictions across augmented versions of test samples. However, the learned prompt via existing TTPT methods are noisy and irrelevant to task-specific distribution.
This raises the question: How can we effectively adapt a pre-trained visual-language model (e.g., CLIP) to a new task by judiciously incorporating domain-specific information during testing?

Recent works in natural language processing (NLP) have shown that in-context learning (ICL)\footnote{Without the need to fine-tune any model parameters for downstream tasks, in-context learning involves adding domain-specific input-output pairs, referred to as in-context examples, to a test example. This prompts the model to learn relevant patterns for the test sample, enabling it to perform multiple tasks seamlessly.}~\cite{brown2020language} is effective in adapting foundation models for downstream tasks including knowledge 
retrieval~\cite{tay2022transformer,wang2020language} and multi-choice tasks~\cite{min2022rethinking}. Leveraging in-context learning, large language models (e.g., ChatGPT) can tackle novel tasks solely through inference, conditioning on a limited set of input-label pairs to make predictions for new inputs. However, the visual model has not yet achieved this in-context learning ability. Existing works~\cite{sun2020test,liu2021ttt} demonstrate that optimizing an unsupervised test-time objective, involving test samples, can enhance model performance on the target data. Inspired by this,
we would like to develop
a novel in-context evaluation
paradigm that improves the performance of existing prompt
learning algorithms in a ChatGPT-like manner. Under this
paradigm, the in-context evaluator operates on test samples at
test-time with unaltered model parameters, where the model
only processes unlabeled test samples and in-context examples.

\begin{figure*}[t]
  \centering
\includegraphics[width=0.92\textwidth]{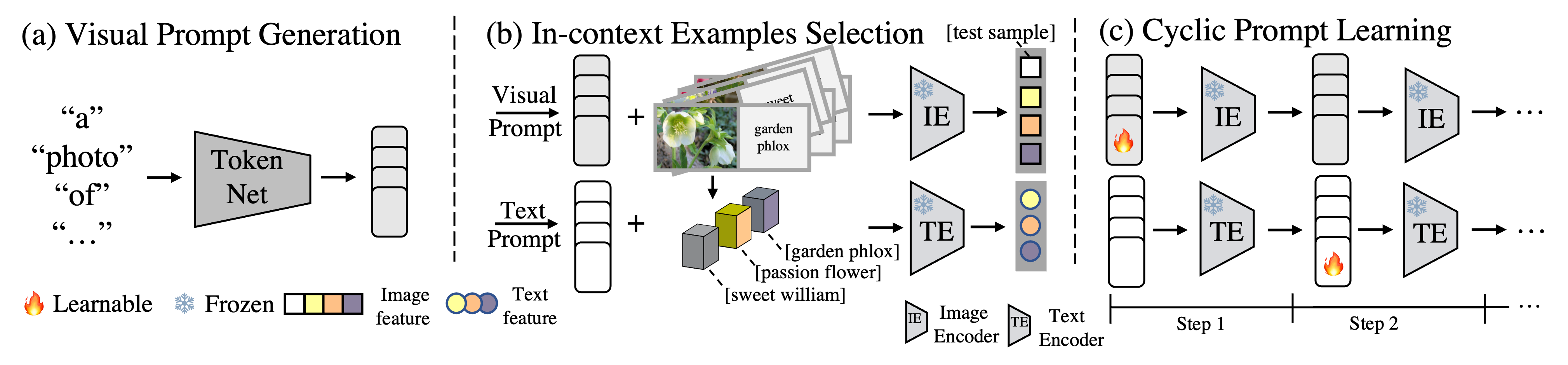}
  \caption{Illustration of the proposed in-context prompt learning (InCPL) framework. (a) We employ a token network to convert language descriptions
into visual prompts, which serve as inputs to the vision encoder of the CLIP model. (b) We construct a test sample coupled with in-context examples and further introduce a context-aware unsupervised loss to optimize visual prompts tailored to test samples.
(c) We design a cyclic learning strategy to seamlessly integrate the visual prompt with the text prompt, which effectively retrieves the pre-trained knowledge relevant to test data across different modalities.}\label{fig:abc}
\end{figure*}

Toward this goal, we propose \textbf{In}-\textbf{C}ontext \textbf{P}rompt \textbf{L}earning (InCPL) that empowers a pre-trained vision-language model, such as the CLIP model, with in-context examples as context information on downstream task. Our approach maintains the frozen weights of the vision-language model while enabling gradient back-propagation through it to learn visual prompt. This enables us to customize visual prompts for different test samples using in-context examples. To achieve this, we undertake the following steps, as illustrated in Figure \ref{fig:abc}:
(\textit{i}) we employ a token network to convert language descriptions
into visual prompts, which serve as inputs to the vision encoder of the CLIP model (Figure~\ref{fig:abc} (a)). (\textit{ii}) We construct a test sample coupled with in-context examples and further introduce a context-aware unsupervised loss to optimize visual prompts tailored to test samples (Figure~\ref{fig:abc} (b)).
(\textit{iii}) We design a cyclic learning strategy to seamlessly integrate the visual prompt with the text prompt, which effectively retrieves the pre-trained knowledge relevant to test data across different modalities (Figure~\ref{fig:abc} (c)). Through these steps, InCPL enables the CLIP model to adapt to new tasks by leveraging domain-specific information, while keeping the model parameters frozen. By harnessing the power of a pre-trained vision-language model, our InCPL inherits the robust zero-shot learning capabilities of the CLIP model for downstream tasks on which it was not initially trained. As a result, our approach can effectively adapt pre-trained model to new tasks with learned visual prompts. Finally, we conduct our experiments on diverse image recognition datasets and evaluate the transfer performance with pre-trained models. Experimental results show that our InCPL outperforms previous methods by large margins, especially on fine-grained datasets. We also provide comprehensive ablation study and qualitative analysis to understand the effectiveness of InCPL.

Our main contributions can be summarized as follows: 1) We propose a straightforward yet highly in-context learning framework to instruct the CLIP model with in-context examples as context information, which enhances
the performance of prompt learning algorithms in a ChatGPT-like manner. 2) We propose an efficient language-to-vision translator to leverage the textual prior information from the language modality, aiding in the learning of visual prompts. 3) We design a cyclic learning strategy for visual 
and textual prompts to ensure mutual synergy across different modalities. 4) Our InCPL approach is rigorously evaluated through extensive experiments, demonstrating superior performance and achieving state-of-the-art (SOTA) results across a diverse set of downstream datasets.

\section{Related work}

\textbf{Prompt Learning in Vision-language Models.} Foundational vision-language (V-L) models~\cite{radford2021learning,jia2021scaling,zhai2022lit} exploit both visual and textual modalities to encode multi-modal representations. These models are pre-trained on a large amount of image-text pairs available online in a self-supervised manner. During the pre-training stage, a contrastive loss function is used to pull together the features of paired images and texts while pushing away unpaired image-text features. V-L models like CLIP~\cite{radford2021learning}, FILIP~\cite{yao2021filip} and Florence~\cite{yuan2021florence} have demonstrated impressive zero- and few-shot generalization capabilities in a wide range of downstream tasks. 
Inspired by prompt learning in natural language processing (NLP), recent works have proposed to adapt V-L models to downstream tasks by learning the prompt tokens. CoOp~\cite{zhou2022learning} prepends a category name with the prompt “a photo of a" (e.g., “a photo of a cat"), 
and optimizes continuous set of prompt vectors at its language branch. CoCoOp~\cite{zhou2022conditional} further makes textual prompt conditioned on each input instance (image) and thus enables dynamic text prompt to adapt to each instance. As a visual prompt technique, VP~\cite{bahng2022exploring,jia2022visual} introduces additional random noise as task-specific learnable parameters into input image space. Their prompts are exclusively concentrated within the visual branch. 
Afterwards, MaPLe~\cite{khattak2023maple} extends CoOp~\cite{zhou2022learning} to effectively leveraging two kinds of prompts to ensure synergy between vision-language modalities during training stage.
However, these learned prompts are limited to the training data distribution and their performances significantly degrade when the model is tested on inputs drawn from a different distribution. To fully delve into the utilization of visual prompts for adapting V-L models, we explore a novel approach based on \textit{in-context} learning~\cite{brown2020language} to strongly encourage V-L model adapt to new tasks quickly by using only few in-context examples.

\textbf{Test-time Adaptation Methods.} Test-time adaptation (TTA)~\cite{goyal2022test} focus on challenging scenarios where only a source model and unlabeled target data are available. This work primarily attempts to address the distribution shift by designing effective test-time objective for test sample. Two similar paradigms are test-time training (TTT) ~\cite{sun2020test,liu2021ttt} and domain adaptation (DA)~\cite{saito2019semi,wu2019joint}. TTT adds a self-supervised proxy task, such as recognizing rotations of an image, to training objective during training stage and computes an optimization objective at test time, while DA requires the utilization of both source and target data for training with a cross-domain loss function. Without source training data, TTA~\cite{shin2022mm} does not require joint training across losses (TTT) or domain adaptation (DA). The above settings usually need to fine-tune specific model parameters, which may discard valuable knowledge provided by the pre-trained model and increase the risk of model overfitting.
TPT~\cite{shu2022test} is the first test-time prompt tuning work to enhance the zero-shot generalization of CLIP model by learning adaptive prompts for each test sample at test time. DiffTPT~\cite{feng2023diverse} extend test-time
prompt tuning by leveraging pre-trained diffusion models to
augment the diversity of test samples. However, TPT~\cite{shu2022test} and DiffTPT~\cite{feng2023diverse} focus only on unlabeled test sample, which leads to the prompted (retrieved) knowledge features from CLIP model deviates from their original meanings within the target distribution.

\textbf{In-context Learning.} In-context learning (ICL) defined by GPT3~\cite{brown2020language} is a new paradigm, where autoregressive language model can perform on-the-fly computational reasoning on unseen tasks~\cite{min2021metaicl}, given prompts and examples serving as context. Following this idea, \cite{wang2023context} and \cite{wang2023images} use task-specific examples to help model understand the underlying task and perform the same task on a new input. \cite{wang2023context} enable in-context learning in
diffusion-based generative models, while \cite{wang2023images} allows the model to rapidly adapt to visual dense prediction task with labeled examples. However, these works only employ image-label pairs as task prompts and train overall model on a wide range of tasks. Unlike existing methods, our approach maintains the pre-trained model parameters and tunes the input prompts to retrieve prior knowledge relevant to the test samples. Consequently, our method is more efficient.

\subsection{Preliminaries and Problem Definition}
\textbf{CLIP Models.} In this paper, we focus on generalizing the pre-trained vision-language model (i.e., CLIP)~\cite{radford2021learning} to downstream tasks in zero-shot manner while keeping the model parameters frozen. Without supervision from human labels, CLIP can directly learn associated visual-text representations from a large amount of image-text pairs. Specifically, CLIP trains a vision encoder $g_{I}(\cdot)$ and a text encoder $g_{T}(\cdot)$ in contrastive learning to align the embeddings and then match image-text pairs. During testing stage, zero-shot recognition is accomplished by comparing image features with the class features synthesized by the text encoder on class names.

\textbf{Test-time Prompt Learning.} With pre-trained CLIP model, test-time prompt learning approaches~\cite{shu2022test}
append learnable prompt tokens at the text encoder during test time, i.e., ``a photo of a" is added to every class name to form ``a photo of a \{class\}". This provides the model very helpful context information about the downstream task. Formally, class label is wrapped within a language prompt, which can be formulated as $\{\bm{t}_{SOS}, \bm{P_\text{\textbf{t}}}, \bm{c},\bm{t}_{EOS}\}$. Here, $\bm{c}$ represents the class label, and $\bm{P_\text{\textbf{t}}}=\{\bm{P_l}\}_{l=1}^{4}$ denotes the learnable prompt initialized with word embeddings` corresponding to ``a photo of a".
$\bm{t}_{SOS}$ and $\bm{t}_{EOS}$ are
the learnable start and end token embeddings.

\section{Method}
\begin{table*}[t]
\centering
  \caption{\small Characteristics of in-context test-time adaptation (ITTA) and other adaptations.}
\scalebox{0.8}{
    \begin{tabular}{l|cc|ccc|cc}
    \toprule
    \multirow{2}[2]{*}{Setting} & \multicolumn{2}{c|}{Adaptation} & \multicolumn{3}{c|}{Available Data} & \multicolumn{2}{c}{Loss}  \\
          & Train & Test  & Source & Target & In-context & Train loss & Test loss  \\
    \midrule
    Fine-tuning &   \checkmark    &$\times$&  -  & $x_{t},y_{t}$  &     -  &   $L(x_{t},y_{t})$    &      - \\
    Domain Adaptation & \checkmark & $\times$ & $x_{s},y_{s}$ & $x_{t}$ &  - & $L(x_{s},y_{s})+L(x_{s},x_{t})$  &  -   \\
    Domain Generalization & \checkmark & $\times$ & $x_{s},y_{s}$ &  & - & $L(x_{s},y_{s})$  & -      \\
    Test-Time Training &  \checkmark & \checkmark & $x_{s},y_{s}$ & $x_{t}$  &  -  & $L(x_{s},y_{s})+L(x_{s})$ &  $L(x_{t})$     \\
    Fully Test-Time Adaptation & $\times$ & \checkmark &  - &  $x_{t}$   & - &  -    &  $L(x_{t})$  \\
    \midrule
    ITTA &  $\times$ & \checkmark  &   -   &   $x_{t}$       &  $x_{i},y_{i}$    & -      &  $L(x_{t})+L(x_{i},y_{i})$  \\
    \bottomrule
    \end{tabular}}
  \label{tab:setting}
\end{table*}

\textbf{Problem Definition.} Given a pre-trained model $f_{\theta_0}$ with parameters $\theta_0$ and prompt $\bm{P}$, the proposed In-context Test-Time Adaptation (ITTA) aims to adapt $f_{\theta_0}$ to downstream task with only few input-label pairs (known as in-context examples) $(x_{i},y_{i})$ and test samples $x_t$. During testing stage, ITTA fine-tunes a learnable prompt using an unsupervised optimization objective where test samples lack labels, but in-context examples are accessible. The candidate in-context example set is formed by selecting one image-label pair from each category, followed by random sampling of several pairs from this set to create in-context examples. For a more comprehensive discussion on the selection of examples, we refer to the Section~\ref{in-context-appendix}. In the absence of source domain data $(x_{s}, y_{s})$, domain adaptation and test-time training, which require joint training across different domains, are not feasible. While test-time adaptation also does not rely on source domain data, our ITTA can provide additional labeled examples as context information for the test domain. The detailed differences between ITTA and other adaptation methods are shown in Table~\ref{tab:setting}.

\subsection{Visual Prompt Input with Images}

\begin{figure}[t]
  \centering
\includegraphics[width=0.7\linewidth]{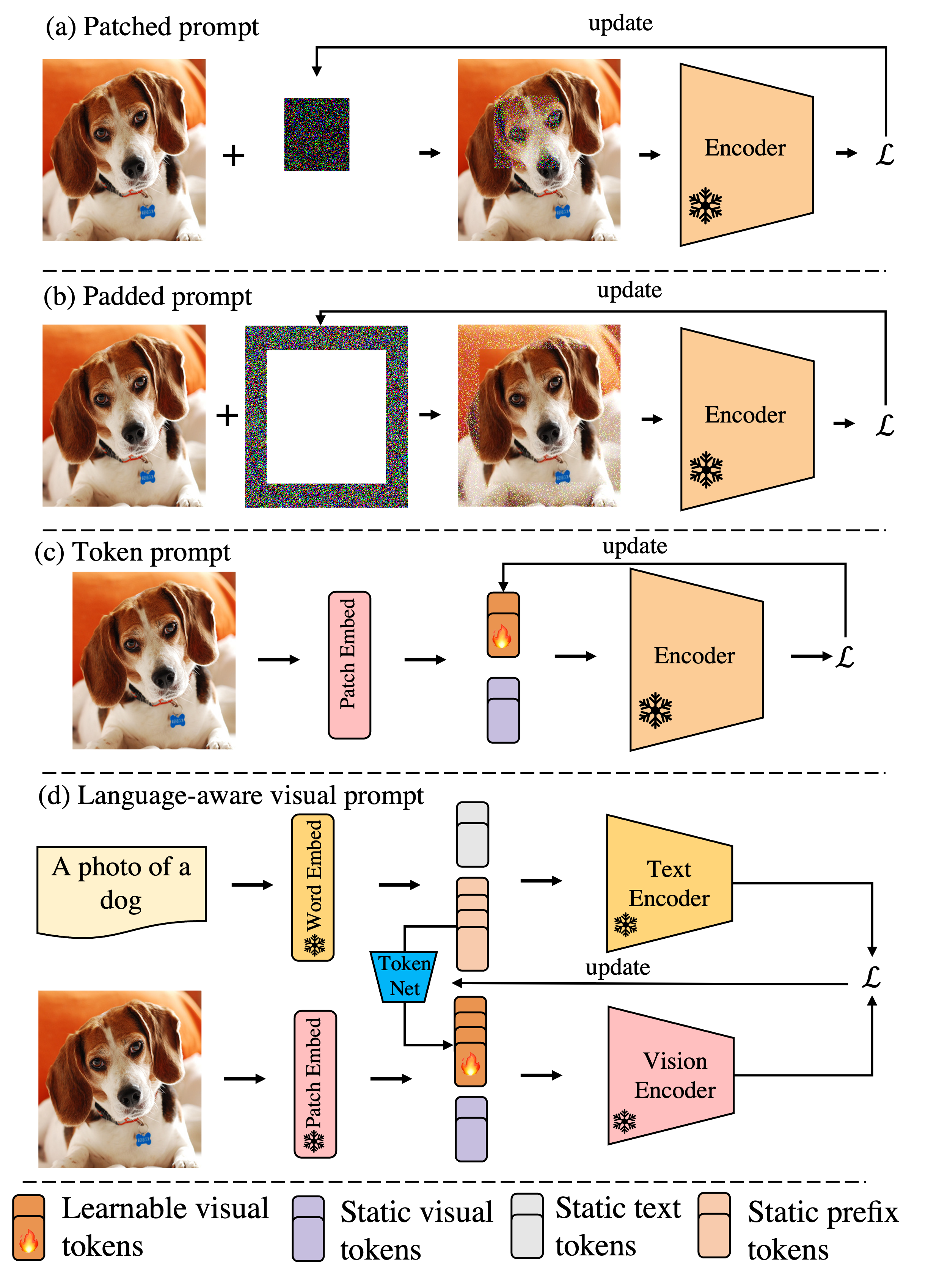}
  \caption{The comparison between our language-aware visual prompt approach and the patched, padded, and token-based visual prompt methods.}
  \label{fig:prompt}
\end{figure}

The visual prompt approach enables the learning of image perturbations (such as noisy boxes or rectangles) for a frozen model. This enables the model, when prompted with these perturbations, to perform a new task. As depicted in Figure~\ref{fig:prompt} (a) and (b), recent studies~\cite{bahng2022exploring,jia2022visual} have introduced task-specific learnable parameters as visual prompts. These prompts are prepended into the input image space and are learned during the fine-tuning stage. However, these parameters are typically initialized with random noise. We argue that a general text description (e.g., a photo of a) provides richer visual context information compared to random noise. 
This is because linguistic words can be perceived as interpretable lexical tokens that can be easily translated into visual content comprehensible to the vision encoder. This is especially advantageous during the test-time adaptation, which lacks effective supervision information.

\begin{figure}[t]
  \centering  
  \includegraphics[width=0.35\textwidth]{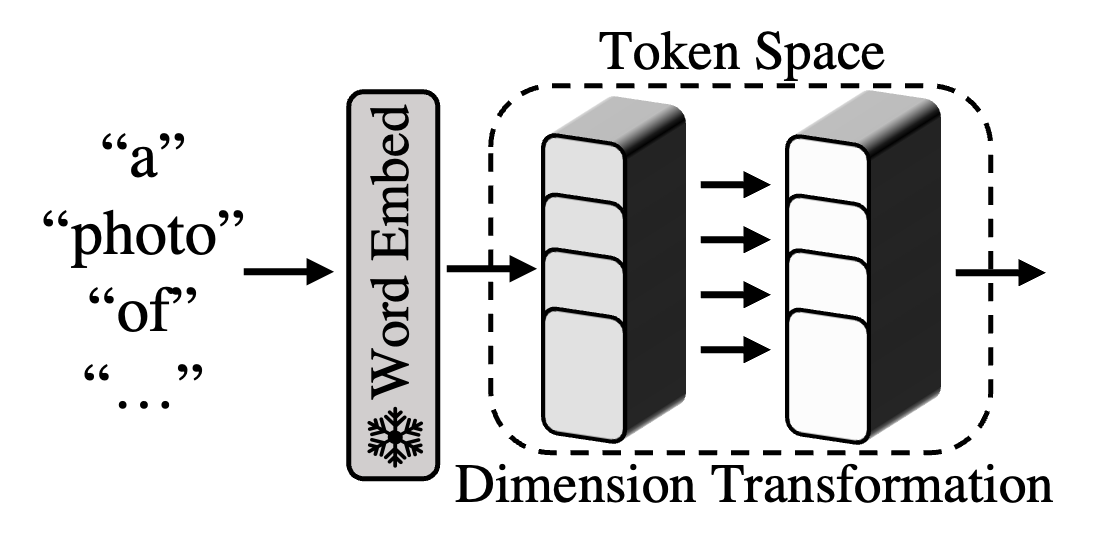}
  \caption{The illustration of token net.}
  \label{fig:token-net}
\end{figure}

A straightforward approach to implementing a visual prompt is to combine patch image embedding with learnable parameter vectors initialized with word embeddings. However, this design not only lacks a seamless connection between the vision and language modalities but also faces the challenge of differences in dimensionality between these two types of embeddings. We propose an efficient language-to-vision translation design to leverage the textual prior information from the language modality, aiding in the learning of visual prompts. Specifically, we introduce a lightweight neural network, known as Token-Net, to
convert textual descriptions into a vision prompt that the vision encoder can comprehend. Token-Net is responsible for generating a conditional token vector for each input image, which is then integrated with the context visual vectors. For a illustration of this architecture, please refer to Figure~\ref{fig:token-net}. 

Formally, let $\mathcal{F}_{\theta}$ denote token net parameterized by $\theta$, each visual prompt $\bm{P}_{\text{\textbf{v}}}$ is obtained
by projecting text words in a prompt $\bm{P}_{\text{\textbf{t}}}$ into a set of learnable vectors via language-to-vision token net, such that $\bm{P}_{\text{\textbf{v}}}=f_{\theta}(\bm{P}_{\text{\textbf{t}}})$. Given as input image $\textbf{X}\in \mathbb{R}^{C \times H \times W}$, we divide it into $M$ pathes and produce patch tokens 
through a projection at vision branch of CLIP model.
The visual prompt for the $i$-th input image is thus conditioned on the language description. The token net is built with a linear layer, which maps inputs of dimensions $d_l$ to $d_v$. We
leave the exploration of more advanced designs to future endeavors. During training, we update the context vectors
$\bm{P}_{\text{\textbf{v}}}$ together with Token-Net's parameters $\mathbf{\theta}$. This works as a bridge between visual and language modalities, thus reducing modality discrepancy and encouraging knowledge transfer from language to vision. 
Unlike CoCoOP~\cite{zhou2022conditional}, which solely conditions a text prompt on each input image without visual prompts, the explicit conditioning of our visual prompt on natural language ensure the initial visual prompts within a meaningful embedding space, thereby facilitating a smoother convergence process.

\begin{figure}[t]
  \centering
\includegraphics[width=0.8\linewidth]{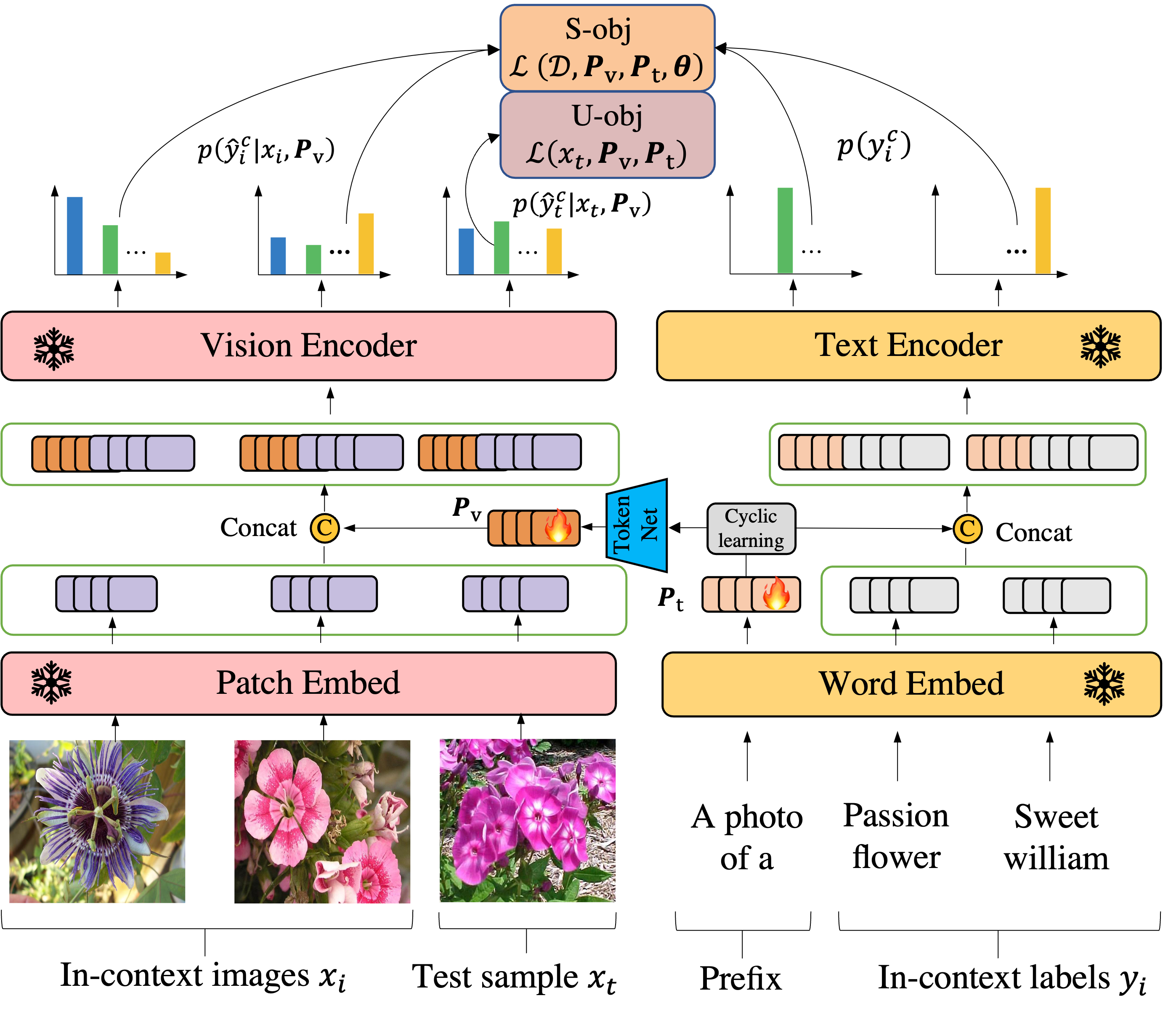}
  \caption{Illustration of the proposed visual in-context prompt learning for test-time visual recognition. Each in-context example ($x_i, y_i$),  test sample $x_t$, and its prefix text are fed into a token encoder to obtain visual, prefix, and text tokens. The text tokens are translated into a visual prompt using a token network. We optimize the visual tokens using a context-aware objective: a supervised cross-entropy term $L(x_i, y_i,\bm{P}_{\text{\textbf{v}}})$ involving in-context examples and an unsupervised entropy minimization $L(x_t,\bm{P}_{\text{\textbf{v}}})$ with the test sample.}
  \label{fig:overview}
\end{figure}

\subsection{In-context Learning for Visual Recognition}

In-context learning is an emerging paradigm that involves supplying labeled examples as input to a pre-trained language model. Alongside these examples, a query example is provided for which the model must predict the label based on its understanding of the task learned from the provided examples. Unlike existing few-shot prompt learning methods, which learn share prompt for all test data, our InCPL can learn adaptive prompts tailored to different test samples. The illustration of the proposed InCPL is shown in Figure \ref{fig:overview}.

Formally, given an unlabeled test sample $x_t$ and some in-context examples $\mathcal{D}=\{x_{i},y_{i}\}_{i=1}^{N}$ containing $N$ image-label pairs (i.e., an image and its class name), in-context learning can be formulated as 
\begin{equation}
y_t=\mathcal{F}(\mathcal{D},x_t;\Phi).
\end{equation}

Here in-context examples $\mathcal{D}$ offer contextual guidance (prompt) to the model $\mathcal{F}(\cdot;\Phi)$, enabling it to generate the optimal label $y_t$ for $x_t$ without updating model parameters. Traditional in-context learning relies heavily on the pre-trained model, while its objective on downstream task may well not align with that used in pre-training. Hence, our InCPL appends 
visual prompts given as $\bm{P}_{\text{\textbf{v}}}$ with the visual tokens $\{ e_{cls}, e_1, e_2, \ldots, e_M\}$, and tune them to learn task-specific information. 
Specifically, the visual encoder processes the following input 
tokens $\widetilde{X}_p = \{\bm{P}_{\text{\textbf{v}}}, e_{cls}, e_1, e_2, \ldots, e_M\}$  to produce the prompted visual feature represented as $f_{\widetilde{p}} = \mathcal{F}(X_p; \Phi) $. Then, visual prompt is learned on the following objective to provide the model with the helpful context information about on-the-fly test sample $x_t$.

\textbf{Test-time Objective.} Given that sample's label is not available for test-time adaptation, we propose a context-aware unsupervised loss for visual in-context prompt learning. At each time step of adaptation, we comprise a batch of test inputs with on-the-fly test sample and in-context examples. For unlabeled test sample $x_t$, our test-time 
unsupervised objective (U-obj) is to minimize the Shannon entropy~\cite{shannon1948mathematical}, $L(x_t,\bm{P_{\text{\textbf{v}}}}) = -\sum_{c} p(\hat{y}^c) \log p(\hat{y}^c|x_{t},\bm{P_{\text{\textbf{v}}}})$ for the model predictions $\hat{y} = f_{\theta}(x_t)$ of class $c$. Given image-label pairs $\mathcal{D}=\{x_{i},y_{i}\}_{i=1}^{N}$, our test-time supervised objective optimize 
the same loss on in-context examples with a supervised loss (S-obj) $L(x_i,y_i,\bm{P_{\text{\textbf{v}}}})= -\sum_{c} p(y_i) \log p(\hat{y}^{c}_{i}|x_{i},\bm{P_{\text{\textbf{v}}}})$. We take \textit{a gradient step} towards optimizing the context-aware unsupervised loss $L=L(x_t,\bm{P_{\text{\textbf{v}}}})+\sum_{i=1}^{N}L(x_i,y_i,\bm{P_{\text{\textbf{v}}}})$ over learnable parameters $\bm{P_{\text{\textbf{v}}}}$ that are shared across the test batch. The overall learning objective is formulated as follows:  
\begin{equation}  
\bm{P_{\text{\textbf{v}}}^{*}}= arg \min_{\bm{P_{\text{\textbf{v}}}}} \left\{L(x_t,\bm{P_{\text{\textbf{v}}}})+{\textstyle \sum_{(x_i,y_i)\in\mathcal{D}}\lambda L(x_i, y_i,\bm{P_{\text{\textbf{v}}}},\theta) }\right\}.
\end{equation}
Here $\lambda$ is loss weight parameter balancing the contributions of different loss components.

\subsection{Cyclic Learning of Visual and Language Prompts}

Visual and language (V-L) prompts are two parameter-effective prompt-tuning methods. Most methods~\cite{khattak2023maple,khattak2023PromptSRC} attempt to simultaneously adapt vision and language branch of CLIP model with learnable prompts of both modalities, which ensures mutual synergy between V-L prompts. This mutual learning strategy performs effectively when training data provides alignment information across visual and language modalities. However, this performance is limited for the target distribution during test-time adaptation, where on-the-fly test sample is unlabeled (i.e., only visual information is available).

To this end, we propose a cyclic
learning strategy for visual and textual prompts to ensure the
mutual synergy across different modalities. Specifically, we start by optimizing one prompt and subsequently utilize the knowledge gained from this step to guide the optimization of the remaining prompt. This process provides the model with richer contextual information, allowing it to thoroughly capture relevant patterns of on-the-fly test sample to align visual and language modalities. The optimization objective for the proposed cyclic learning strategy can be formulated by

\begin{equation} \scriptsize
\mathcal{L}(x_t, \mathcal{D}, P_v, P_t, \theta) = 
\begin{cases}
\begin{aligned}
\arg \min_{\bm{P}_{\textbf{v}}, \theta} \Bigg\{ & L(x_t, \bm{P}_{\textbf{v}}) + \sum_{(x_i, y_i) \in \mathcal{D}} \lambda L(x_i, y_i, \bm{P}_{\textbf{v}}, \bm{P}_{\textbf{t}}, \theta) \Bigg\}, \\
& \quad \text{if } s = 1,
\end{aligned} \\
\begin{aligned}
\arg \min_{\bm{P}_{\textbf{t}}} \Bigg\{ & L(x_t, \bm{P}_{\textbf{t}}) + \sum_{(x_i, y_i) \in \mathcal{D}} \lambda L(x_i, y_i, P_v, \bm{P}_{\textbf{t}}, \theta) \Bigg\}, \\
& \quad \text{if } s = 2.
\end{aligned}
\end{cases}
\end{equation}

where $x_t$ is unlabeled test sample, $\mathcal{D}$ contains in-context examples, $\theta$ is the parameter of token net, and $s=1,2$ are the step numbers of model optimization for visual and text prompt, respectively. 
It is noted that the paragraphs above section 3.4 are focused on visual prompt learning, which is a single-step optimization process. In this section, we introduce text and visual prompt learning in cyclic strategy, which is 
is a two-step optimization process. Two kinds of prompts are optimized in sequence, i.e., first visual then textual prompts, forming the ``cyclic prompt learning" concept.

\subsection{Discussion}
In-context learning allows large language models (GPT3~\cite{brown2020language}, Llama~\cite{touvron2023llama}) to perform inference on unseen tasks by conditioning on in-context examples (a.k.a. prompt) without updating the model parameters. Inspired by this, existing works~\cite{wang2023context,zhang2023makes,wang2023images} explore ``in-context learning" concept for vision model, in which the model is updated using in-context examples.
Meanwhile, CLIP itself is not able to conduct in-context learning task. To equip CLIP with this ability, our InCPL introduces learnable prompt for each test sample in test-time stage. In this way, the model can automatically understand the underlying task with in-context examples.

\textbf{Comparison with Few-shot Learning.} As few-shot methods, CoOP~\cite{zhou2022learning} and CoCoOP~\cite{zhou2022conditional} fine-tune the prompt on ImageNet dataset using 16-shot training data per category and evaluate the generalization performance on downstream tasks. Our work primarily differs from few-shot methods in two main aspects, i.e., sample selection and quantity. (a) For sample selection, few-shot uses strict categories with specific number of samples, which are widely used in training stage.
Differently, in-context learning has no constraint on category.
The in-context samples in testing stage can either share the same category as the current test sample or the irrelevant category. It is also impractical to know the exact category of unlabeled test sample in advance. 
(b) For sample quantity, few-shot learning requires a predefined number of samples from each category, while in-context learning uses a small, arbitrary set of labeled samples-commonly just five samples.

\textbf{Comparison with Semi-supervised Learning.} Semi-supervised learning~\cite{tarvainen2017mean,sohn2020fixmatch} typically incorporates labeled data during the training phase, amalgamating it with unlabeled data to fine-tune the model and improve its performance on unlabeled samples. Labeled data in semi-supervised learning often shares categories with the unlabeled data. In our method, there is no inherent relationship between in-context examples (labeled data) and the test sample, as they are both drawn from the same domain dataset. Our approach does not necessitate any category information about the test sample, distinguishing it from semi-supervised learning methods.

\section{Experiments}

\subsection{Datasets and Implementation details} \label{data implement}
\subsubsection{Datasets}
We assess the transfer performance of our method across 10 fine-grained classification datasets and 4 out-of-distribution (OOD) datasets from ImageNet. Since our method primarily focuses on test-time model adaptation, our evaluation is exclusively based on the testing dataset across all these datasets.

\textbullet~\textbf{Flower102}~\cite{nilsback2008automated} is a widely used dataset consisting of 102 different categories of flowers. Each category consists of between 40 and 258 images. It is commonly employed for fine-grained image classification tasks.

\textbullet~\textbf{OxfordPets}~\cite{parkhi2012cats} is a dataset designed for pet image classification, containing a large variation in scale, pose, and lighting conditions. It contains images of 37 different pet breeds with roughly 200 images for each class.

\textbullet~\textbf{Food101}~\cite{bossard2014food} is a dataset specifically curated for food recognition applications. It contains images of 101 different food categories, making it suitable for tasks related to food image classification and analysis.

\textbullet~\textbf{Describable Textures Dataset (DTD)}~\cite{cimpoi2014describing} is a 
texture dataset, which consists of 5640 images. These images are classified into 47 distinct categories, inspired by human perception, with precisely 120 images allocated to each category.

\textbullet~\textbf{StanfordCars}~\cite{krause20133d} is a dataset commonly used for fine-grained car classification tasks. This dataset contains 16,185 images of 196 classes of cars, which is split into 8,144 training images and 8,041 testing images.

\textbullet~\textbf{Aircraft}~\cite{maji2013fine} dataset contains 10,200 images of various aircraft, with 100 images for each of 102 different aircraft model variants, most of which belong to airplanes.

\textbullet~\textbf{UCF101}~\cite{soomro2012ucf101} is a widely recognized dataset for human action recognition, which consists of 13,320 video clips spanning 101 different human action categories. These 101 categories are further  classified into 5 types, including Body motion, Human-object interactions, Playing musical instruments, Sports, and Human-human interactions.

\textbullet~\textbf{EuroSAT}~\cite{helber2019eurosat} is a dataset and deep learning benchmark designed for land use and land cover classification. It is based on Sentinel-2 satellite images with 13 spectral bands and a total of 27,000 labeled and geo-referenced images with 10 distinct classes.

\textbullet~\textbf{Caltech101}~\cite{fei2004learning} dataset is composed of approximately 9,000 images with 101 object categories and a background category. Each object category contains approximately 40 to 800 images, with typical image sizes of 200-300 pixels.

\textbullet~\textbf{SUN397} dataset encompasses 108,753 images spanning 397 categories, serves as a benchmark in scene understanding studies. Each category in this diverse collection is represented by a minimum of 100 images.

\textbullet~\textbf{ImageNet}~\cite{deng2009imagenet} dataset is a large-scale ontology of images built upon the backbone of the WordNet structure, designed to advance the field of computer vision. It spans 1000 distinct object classes and contains 1,281,167 training images, 50,000 validation images and 100,000 test images. 

\textbullet~\textbf{ImageNet-V2}~\cite{recht2019imagenet} is a test set including natural images collected from various sources. This dataset is consisted of 10,000 images distributed across 1,000 distinct ImageNet categories.

\textbullet~\textbf{ImageNet-Sketch}~\cite{wang2019learning} dataset consists of 50000 images, 50 images for each of the 1000 ImageNet classes. These images are obtained by performing Google Image searches using the query "sketch of [standard class name]."

\textbullet~\textbf{ImageNet-A}~\cite{hendrycks2021natural} is a challenging dataset containing real-world, unmodified, and naturally occurring examples that are misclassified by ResNet models. It includes 7,500 intentionally altered and corrupted images with 1,000 categories to assess the robustness of different models.

\textbullet~\textbf{ImageNet-R}~\cite{hendrycks2021many} collects images of ImageNet categories presented in artistic renditions, including a total of 30,000 images across 200 distinct ImageNet categories.

\textbf{Implementation Details.} We apply in-context prompt learning on a pre-trained ViT-B/16 CLIP model. We minimize the semi-parametric objective loss to optimize the visual prompt for 1 step and cyclically alternate between optimizing the visual and prompt for 2 steps. All models are trained with in-context examples and test sample on a single NVIDIA GPU. We use the AdamW~\cite{loshchilov2017decoupled} optimizer with a learning rate of $5e^{-3}$ for all datasets. By default, we set weight parameter $\lambda=0.4$. If not specifically emphasized, we adopt in-context prompt learning with only visual prompt as our default setting for all ablation studies. 
We select 5 in-context examples from a fixed subset of labeled data, which is composed of randomly sampling 1 sample from each category. The context samples only provide the task information to do on the test sample.
They are usually from the same target dataset, but there are no other relationships between them, such as category. Token net is randomly initialized at the start of test-time adaptation and accumulatively updated across the entire evaluation process. For each test sample, $P_{\text{t}}$ is initialized with prefix tokens derived from ``a photo of a", which is then converted into a visual token. 
$P_{\text{v}}$ is then initialized by the above learned $P_{\text{t}}$.

\begin{table*}[t]
  \centering
  \caption{Accuracy comparison with previous methods on fine-grained classification datasets. CoOp and CoCoOp are fine-tuned on the ImageNet dataset using 16-shot training data per category. Baseline CLIP, prompt ensemble, and TPT do not require ImageNet dataset as training data. Our method builds above TPT and further learns sample-specific prompts for test sample by using only few in-context examples. The top-1 classification accuracy is reported on each dataset.}
  \scalebox{0.73}{
    \begin{tabular}{llccccccccccc}
    \toprule
    Method & Type  & Flower & DTD   & Pets  & Cars  & UCF101 & Caltech & Food & Aircraft & EuroSAT & SUN & Average \\
    \midrule
    CLIP-ViT-B/16 & Zero-shot & 67.44 & 44.27 & 88.25 & 65.48 & 65.13 & 93.35 & 83.65 & 23.67 & 42.01 & 62.59 & 63.58 \\
    \midrule
    Ensemble~\cite{shu2022test} & Zero-shot & 66.99 & 45.04 & 86.92 & 66.11 & 65.16 & 93.55 & 82.86 & 23.22 & 50.42 & 65.63 & 64.59 \\
     TPT~\cite{shu2022test}   & Zero-shot & 68.98 & \textbf{47.75} & 87.79 & 66.87 & 68.04 & 94.16 & \textbf{84.67} & 24.78 & 42.44 & 65.50 & 65.10 \\
     DiffTPT~\cite{feng2023diverse}   & Zero-shot & 70.10 & 47.00 & 88.22 & 67.01 & 68.22 & 92.49 & \textbf{87.63} & 25.60 & 43.14 & 65.74 & 65.51 \\
    CoOp~\cite{zhou2022learning}  & Few-shot & 68.71 & 41.92 & 89.14 & 64.51 & 66.55 & 93.70  & 85.30  & 18.47 & 46.39 & 64.15 & 63.88 \\
    CoCoOp~\cite{zhou2022conditional}  & Few-shot & 70.85 & 45.45 & 90.46 & 64.90  & 68.44 & 93.79 & 83.97 & 22.29 & 39.23 & 66.89 & 64.63 \\
    \midrule
    InCPL & In-context & \textbf{72.27} & 47.58 & \textbf{90.62} & \textbf{67.54} & \textbf{70.26} & \textbf{94.69} & 84.62 & \textbf{24.99} & \textbf{64.52} &\textbf{ 67.93}& \textbf{68.50} \\
    \midrule
    \end{tabular}}
  \label{tab:fine-grained}
\end{table*}

\subsection{Generalize CLIP to Fine-grained Classification Datasets}  

In this experiment, we investigate the cross-dataset generalization ability of CLIP model from ImageNet to various fine-grained datasets. Experimental results are reported in table~\ref{tab:fine-grained}.  It can be observed that all prompt learning approaches perform better than the zero-shot generalization of CLIP model. However, the learned few-shot prompts (i.e., CoOp~\cite{zhou2022learning}, CoCoOp~\cite{zhou2022conditional}) are constrained by the distribution corresponding to training data and may exhibit limited generalization beyond that, especially for fine-grained datasets. 
The learned zero-shot prompts (e.g., TPT~\cite{shu2022test}) catastrophically underfit to novel tasks, whereas our model successfully learns all tasks with the help of in-context information. As shown in Table~\ref{tab:fine-grained}, our model 
achieves the best performance over zero-shot and few-shot baselines in terms of average accuracy.

\begin{table}[t]
  \centering
  \caption{Comparison with existing methods across a range of distribution shifts on the ImageNet dataset. CoOp,  CoCoOp, CLIPOOD, MaPLe, and PromptSRC are trained on source domain (ImageNet) and evaluated on out-of-domain datasets (ImageNet variant distributions). Baseline CLIP, prompt ensemble, and TPT don't require ImageNet dataset as training data.}
  \scalebox{0.7}{
    \begin{tabular}{llcccccc}
    \toprule
    \multirow{2}[2]{*}{Method} & \multicolumn{1}{l}{\multirow{2}[2]{*}{Type}} & ImageNet & ImageNet-A & ImageNet-V2 & ImageNet-R & ImageNet-S & \multirow{2}[2]{*}{Average} \\
          &       & Top 1 acc. & Top 1 acc. & Top 1 acc. & Top 1 acc. & Top 1 acc. &  \\
    \midrule
    CLIP-ViT-B/16 & Zero-shot & 66.73  & 47.87 & 60.86 & 73.98 & 46.09 & 59.11 \\
    \midrule
    Ensemble~\cite{shu2022test} & Zero-shot & 68.34 & 49.89 & 61.88 & 77.65 & 48.24 & 61.20 \\
    TPT~\cite{shu2022test}   &  Zero-shot   & 68.98  & 54.77 & 63.45 & 77.06 & 47.94 & 61.44 \\
    CoOp~\cite{zhou2022learning}  & Few-shot & 71.51 & 49.71 & 64.20  & 75.21 & 47.99 & 61.72 \\
    CoCoOp~\cite{zhou2022conditional} & Few-shot & 71.02& 50.63 & 64.07 & 76.18 & 48.75 & 62.13 \\
    CLIPOOD~\cite{shu2023CLIPood} & Few-shot & 71.60 & 50.4  & \textbf{64.9}  & 77.2  & --   & -- \\
    MaPLe~\cite{khattak2023maple} & Few-shot & 70.72 & 50.90  & 64.07 & 76.98 & 49.15 & 62.36 \\
    PromptSRC~\cite{khattak2023PromptSRC} & Few-shot &71.27  & 50.90  & 64.35 & \textbf{77.80}  & \textbf{49.55} & 62.77 \\
    \midrule
    InCPL & In-context & \textbf{71.62} & \textbf{56.51}  & 63.87&   77.63 &  48.08 & \textbf{63.54} \\
    \bottomrule
    \end{tabular}}
  \label{tab:ood}%
\end{table}%

\subsection{Generalize CLIP to Distribution Shift} 
In this experiment, we evaluate the out-of-distribution generalization of
CLIP model on different variants of ImageNet. A particular
in-distribution dataset is ImageNet, while we use several out-of-distribution (OOD) datasets from ImageNet variants, i.e., ImageNet-V2~\cite{recht2019imagenet}, ImageNet-Sketch~\cite{wang2019learning}, ImageNet-A~\cite{hendrycks2021natural}, and
ImageNet-R~\cite{hendrycks2021many}.  The performance on four variants of ImageNet with distribution shifts is summarized in Table \ref{tab:ood}. 
As shown in Table~\ref{tab:ood}, CoOp~\cite{zhou2022learning}, CoCoOp~\cite{zhou2022conditional}, CLIPOOD~\cite{shu2023CLIPood}, MaPLe~\cite{khattak2023maple}, and PromptSRC~\cite{khattak2023PromptSRC} are few-shot prompt tuning methods, which are trained on ImageNet using 16-shot training data per category. This is a type of rapid adaptation from ImageNet to ImageNet variants, thus achieving better classification accuracies than zero-shot generalization of CLIP model. TPT~\cite{shu2022test} learns the language prompt using only the given test sample and enhances the zero-shot performance of CLIP model without any task-specific training data. Our InCPL extends TPT to visual prompt learning and learns task-specific information from in-context examples of testing data distribution. 
While our InCPL is not trained on ImageNet dataset, it inherits the zero-shot 
abilities of CLIP model and achieves on-par generalization as ImageNet trained baselines (CoOp~\cite{zhou2022learning}, MaPLe~\cite{khattak2023maple}, PromptSRC~\cite{khattak2023PromptSRC}), surpassing existing methods in terms of average accuracy.

\begin{table}[t]
  \centering
  \caption{Ablation studies on task and sample-specific adaptations in our InCPL. w/ U/S/CU-obj: with unsupervised/supervised/context-aware unsupervised objective.}
  \scalebox{0.7}{
    \begin{tabular}{lcccccc}
    \toprule
    \multirow{2}[2]{*}{Method}& \multirow{2}[2]{*}{Adaptation}& Flower & DTD   & Pets  & Imagenet-R & ImageNet-S \\
    & & Top 1 acc. & Top 1 acc. & Top 1 acc. & Top 1 acc. & Top 1 acc. \\
    \midrule
    Ours w/ S-obj & Task-specific  & 69.36 & 42.43 & 87.84 & 76.89 & 47.24 \\
    Ours w/ U-obj  & Sample-specific & 67.24 & 24.05 & 87.11 & 75.73 & 44.51 \\
    Ours w/ CU-obj & In-context &\textbf{71.13} & \textbf{47.34} & \textbf{90.60}   & \textbf{77.56} & \textbf{48.03} \\
    \bottomrule
    \end{tabular}}
\label{tab:task and instant adaptation}%
\end{table}%

\subsection{Ablation Studies}
\textbf{Component Analysis}. In this experiment, we assess the impact of each component within our method on both task and instance adaptation. With only a limited number of in-context examples, task-specific adaptation employs them as training data to fine-tune a task-aware prompt, which can then be applied to any test sample from a new task. In contrast, sample-specific adaptation does not depend on in-context examples; instead, it learns a sample-aware prompt using only unlabeled test samples.
To investigate the effect of task- and sample-specific adaptation,
We create three distinct variants of our method: 
\textbf{(1) Ours with unsupervised objective (w/ U-obj)} utilize on-the-fly unlabeled test sample to learn sample-specific patterns; \textbf{(2) Ours without supervised objective (w/ S-obj)} adopts 
task-specific labeled examples for each task in prompt learning; 
\textbf{(3) Ours with context-aware unsupervised objective (CU-obj)} adopts labeled examples as task-specific context information and employs test sample to query the prior knowledge of pre-trained model to predict its label.

All the ablation studies are conducted over fined-grained datasets and OOD ImageNet datasets.
As shown in Table~\ref{tab:task and instant adaptation}, our InCPL (w/ CU-obj) significantly surpass other baselines, demonstrating that
test-time adaptation can be improved by introducing task-specific context information with labeled examples. Besides, task- and sample-specific designs in
InCPL can complement each other as InCPL (i.e., the combination of the two designs) clearly outperforms either task-specific adaptation or sample-specific adaptation on the challenging fine-grained
dataset. Moreover, we extend our ablation studies to the
OOD benchmark, and results on ImageNet-R dataset show that the task-specific adaptation achieved 76.86\% accuracy, the sample-specific adaptation
75.73\%, while their combination yielded a 77.56\% accuracy, thereby highlighting the significance of each kind of
adaptation in enhancing InCPL’s performance.

\begin{table}[t]
  \centering
  \caption{ Accuracy of different visual prompt methods.}
  \scalebox{0.7}{
    \begin{tabular}{lccccc}
    \toprule
    \multirow{2}[4]{*}{Method} & Flower & DTD   & Pets  & Imagenet-R & ImageNet-S \\
\cmidrule{2-6}          & Top 1 acc. & Top 1 acc. & Top 1 acc. & Top 1 acc. & Top 1 acc. \\
    \midrule
    Patched prompt & 57.90  & 32.51 & 77.89 & 70.62 & 43.69 \\
    Padded prompt & 56.07 & 32.68 & 79.39 & 68.54 & 40.48 \\
    Token prompt & 59.32 & 34.63 & 74.19 & 70.39	 & 40.1 \\
    Generic-language prompt & 63.30  & 44.74 & 85.2  & 76.46 & 45.29 \\
    Ours (Language-aware) & 71.13 & 47.34 & 90.25 & 77.56 & 48.03 \\
    \bottomrule
    \end{tabular}}
  \label{tab:visual prompt comparison}%
\end{table}

\begin{table}[t] 
  \centering
  \caption{Comparison of different prompt learning strategies with in-context examples and one-shot scenarios. V-P: vision-prompt, T-P: Text-prompt, Con-P: Concurrent-prompt, Cyc-P: Cyclic-prompt. In-con: In-context examples.}
  \scalebox{0.85}{
    \begin{tabular}{llccc}
    \toprule
    \multirow{2}[2]{*}{Method} & \multirow{2}[2]{*}{Type}  & Flower & DTD   & Pets  \\
          &   &     \multicolumn{1}{c}{Top 1 acc.} & \multicolumn{1}{c}{Top 1 acc.} & \multicolumn{1}{c}{Top 1 acc.} \\ 
    \midrule
    CLIP-ViT-B/16 & zero-shot & 67.44 & 44.27 & 88.25 \\
    \midrule
    InCPL w/ V-P & In-Con & 71.13 & 47.34 & 90.60 \\
    InCPL w/ T-P & In-Con & 69.10  & 45.45 & 88.39 \\
    InCPL w/ Con-P & In-Con & 69.71  & 46.04 & 88.50 \\
    InCPL w/ Cyc-P & In-Con & \textbf{72.27}  & \textbf{47.58} & \textbf{90.62} \\
    \midrule
    InCPL w/ V-P & One-shot & 71.99 & 46.57 & 90.35 \\
    InCPL w/ T-P & One-shot& 76.65 & 50.77 & 90.81 \\
    InCPL w/ Con-P & One-shot& 76.09 & 51.60  & 92.04 \\
    InCPL w/ Cyc-P & One-shot & \textbf{82.62} & \textbf{53.84} & \textbf{95.86} \\
    \bottomrule
    \end{tabular}}
  \label{tab:effect on examples}%
\end{table}

\textbf{Comparison with Previous Visual Prompt Approaches.}
Apart from our proposed language-aware visual prompting, we also evaluate different visual prompting methods, including pixel-level patched, padded prompts~\cite{bahng2022exploring,jia2022visual} and token prompt with random initialization. As shown in Figure~\ref{tab:visual prompt comparison}, we learn a single image perturbation by adding learnable patched and padded prompt to input image, which are respectively denoted as ``patched prompt" and ``padded prompt".
Following \cite{jia2022visual}, we also prepend some learnable parameters into input sequence of transformer, denoted as ``token prompt". Table~\ref{tab:visual prompt comparison} shows the accuracies with different prompt designs. Compared with patched-prompt, padded-prompt, and token-prompt, our proposed visual prompt provides a robust language-aware prompt initialization for model optimization, which not only enhances prompt-based performance but also contributes to increase training stability. To further examine the impact of parameter initialization, we experiment with generic-language prompt, which is randomly initialized with a generic context. 
As shown in Table \ref{tab:visual prompt comparison}, generic-language prompt show inferior performance to our language-aware prompt but outperforms the language-unaware prompt methods (i.e., padded, patched and token prompts). This demonstrates that incorporating language modality information into visual branch is beneficial to prompt-based visual recognition.

\textbf{Comparison with Different Prompt Learning Strategies under In-context Examples and One-shot Scenarios.} In our InCPL, in-context examples do not belong to the same categories as the unlabeled test samples. Consequently, they provide task-specific cross-modality alignment information rather than class-specific information, as is typical in traditional few-shot learning. As a comparison, we also use one-shot samples to provide class-specific information during test-time prompt learning.

As shown in Table~\ref{tab:effect on examples}, visual prompt learning significantly improves the zero-shot performance of CLIP compared to language prompt learning. We also evaluate the result of our InCPL when visual and language prompts are concurrently tuned, referred to as ``concurrent-prompt", which yields inferior performance compared to single visual prompt learning. We argue that this is mainly because in-context examples only provide context information and do not offer cross-modal alignment information about the test samples. To further investigate this, we further consider an image-label pair from the same categories with test sample and conduct one-shot learning
for our approach. 
The one-shot can provide class-specific cross-modality alignment information for unlabeled test sample. Compared to in-context learning, one-shot learning significantly improves the performance of InCPL w/ concurrent-prompt. As shown in Table~\ref{tab:effect on examples}, the proposed cyclic prompt learning achieve the best performance and outperform concurrent-prompt with large margin (e.g., +6.53 for Flower dataset). This can be attributed to the disentanglement of contextual information into different modalities through cyclic learning, enabling the model to better utilize labeled data. Here we perform a single iteration of visual and language prompt with sequential learning, and more iterations are expected to yield improved results. We leave the exploration of iterative training for future research.

\begin{figure}[t]
  \centering
\includegraphics[width=0.8\linewidth]{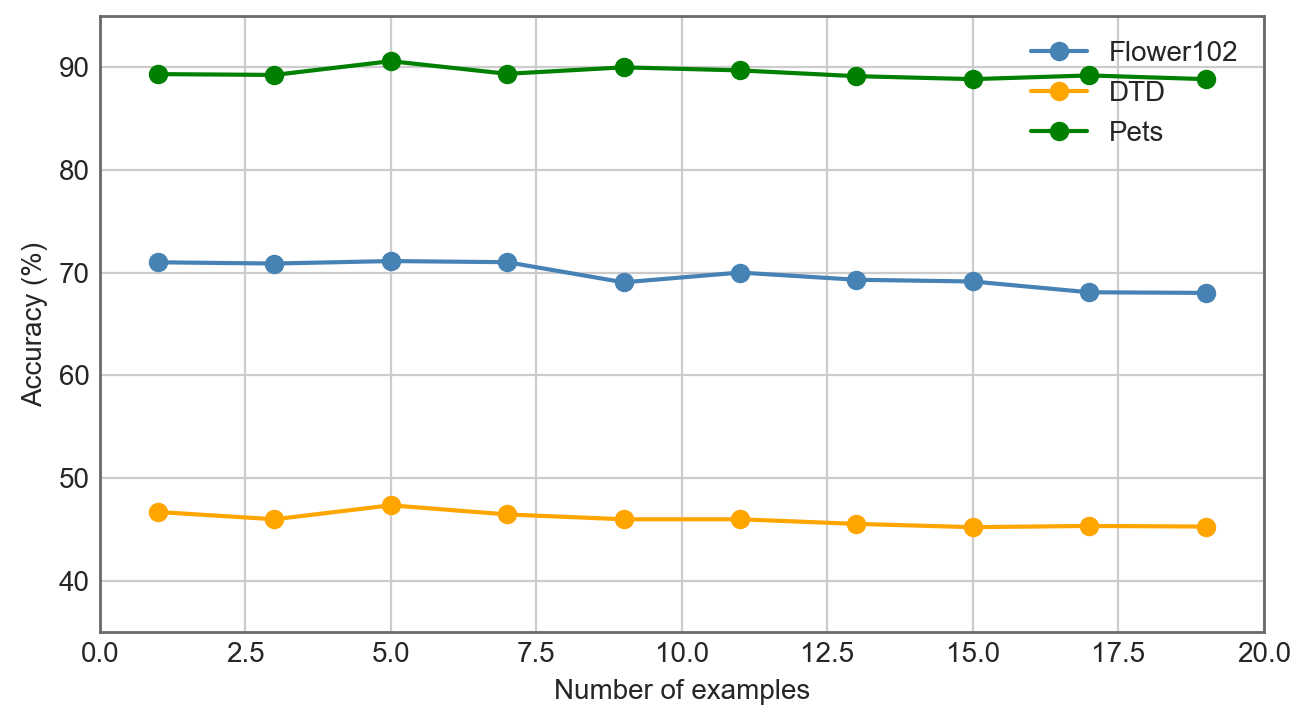}
  \caption{Experimental results w.r.t varied number of in-context examples.}
  \label{fig:num-accuracy}
\end{figure}

\textbf{Does the Number of In-context Examples Matter?} Recall that in-context examples are sampled from the demonstration pool, which contains one sample from each category. We are interested in understanding whether the quantity of these examples has important impacts on model performance. To investigate this, we varied the number of in-context examples from 1 to 19, resulting in a set of results illustrated in Figure~\ref{fig:num-accuracy}. Firstly, it is evident that using in-context examples significantly outperforms the "no examples" method, and the performance clearly benefits from a larger number of examples. However, interestingly, increasing the number of examples beyond a certain point (i.e., 5) starts to decrease performance. This phenomenon may be attributed to the fact that in-context examples can be considered as a form of ``training data," and an abundance of training data may lead the model to learn prompts specific to the in-context examples rather than the unlabeled test samples. While having more examples may not necessarily be advantageous, the crucial question lies in how to strategically select and employ these examples to harness the full potential of the CLIP model. For a comprehensive exploration and discussion of this topic, please refer to Section~\ref{in-context-appendix}.

\textbf{Ground Truth Matters.} To study the impact of correctly-paired inputs and labels in the in-context examples, referred to as ``ground truth input-label mapping", we evaluate the following three methods. 

\begin{figure}[t]
  \centering
\includegraphics[width=1.0\linewidth]{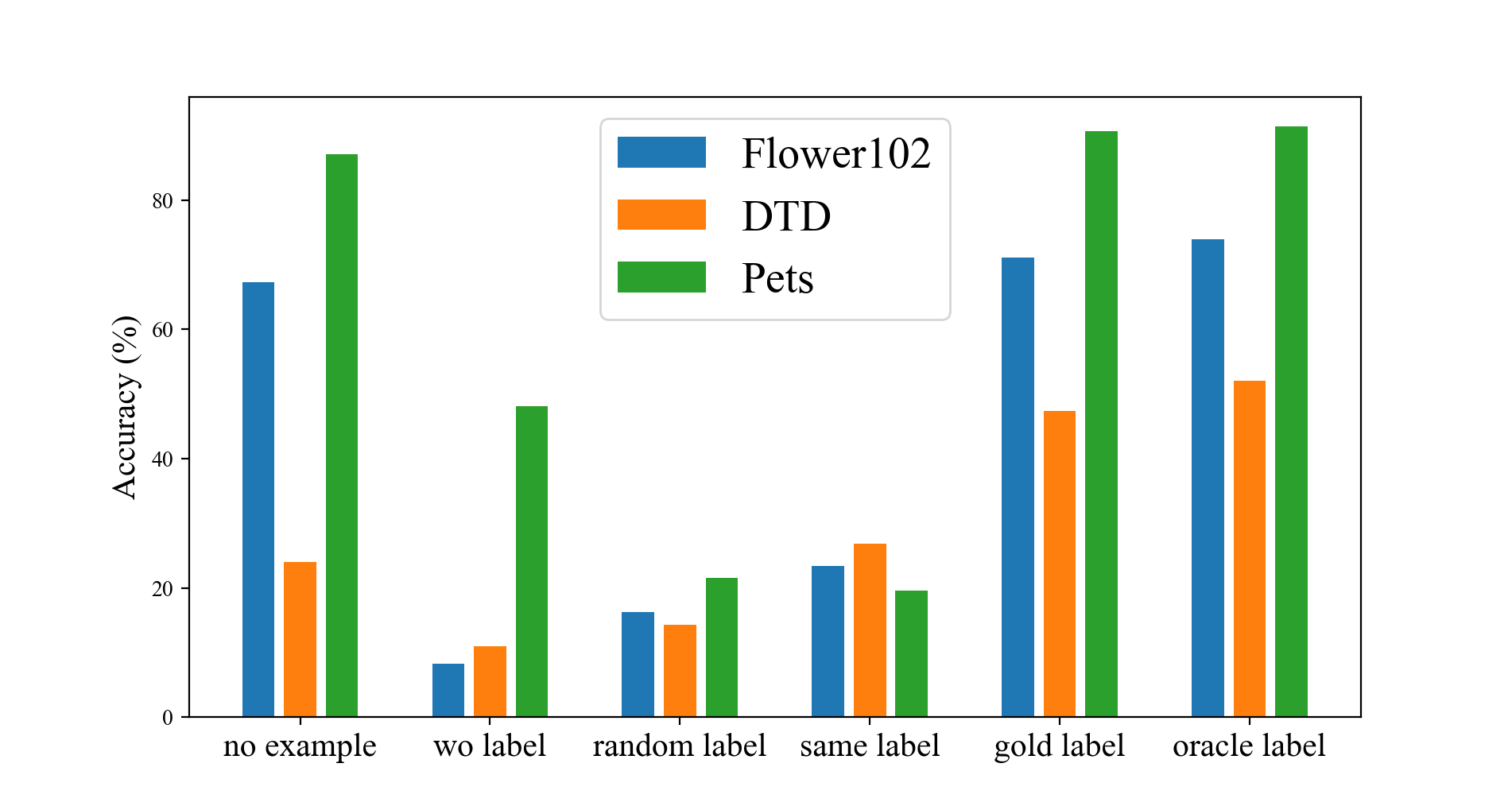}
  \caption{Results when using no-examples, examples w/o labels, examples w/ random labels, examples w/ same labels, examples w/ gold labels and examples w/ oracle labels in fine-grained classification task.}
  \label{fig:label-accuracy}
\end{figure}

\textbullet~\textbf{No example} is a typical test-time adaptation method that does not use any labeled data. A prediction is made by optimizing an unsupervised objective involving the unlabeled test sample.

\textbullet~\textbf{Examples w/o labels} is the baseline that only uses input images without the corresponding labels, indicating the model's performance without considering label information.

\textbullet~\textbf{Examples w/ gold labels} are employed in a typical in-context learning method with a set of labeled examples, indicating the model's performance with access to relevant knowledge.

\textbullet~\textbf{Examples w/ random labels} replace all gold labels with random labels,
which are randomly sampled uniformly from the label space of the testing data.

\textbullet~\textbf{Examples w/ same labels} replace all gold labels with the same label, which is consistent with the label of test sample.

\textbullet~\textbf{Examples w/ oracle labels} replace traditional in-context examples with the oracle examples, where labels are consistent with the label of test sample.

In Fig. \ref{fig:label-accuracy}, we present the recognition accuracy values obtained with different input-label strategies applied to in-context examples. It is evident that model performance is notably sensitive to the correctness of labels. Specifically, using correct gold labels yields better results than employing random labels. Furthermore, employing the same labels as the test sample leads to a significant improvement compared to using random labels. This observation indicates that having consistent labels provides the model with instructive information about the test samples. Moreover, employing examples with oracle labels achieves the upper bound in performance and outperforms the use of examples with gold labels. We attribute this phenomenon to the increased instructive information, which arises not only from the example labels but also from the example inputs themselves. These results underscore the substantial impact of in-context examples' labels on In-Context Learning (ICL) performance, aligning with the findings~\cite{wuself}.

\begin{figure}[t]
  \centering
\includegraphics[width=0.8\linewidth]{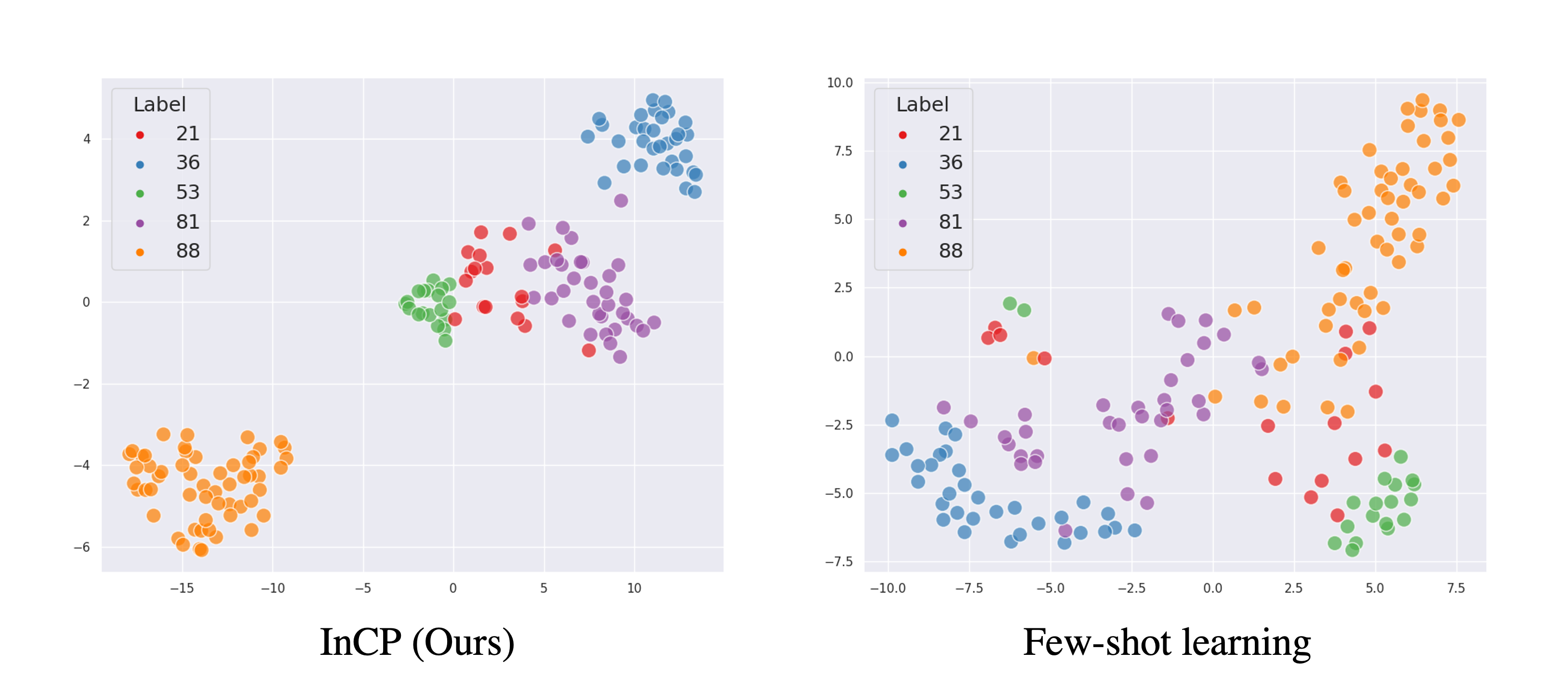}
  \caption{Comparison with few shot learning when using few in-context examples on Flower102.}
  \label{fig:tsne}
\end{figure}

\begin{table}[t]
  \centering
  \small
  \caption{Comparison with CoOp and CoCoOp using the same examples.}
  \scalebox{0.7}{
    \begin{tabular}{lccccc}
    \toprule
  \multirow{2}[3]{*}{Method} & Flower102 & DTD   & Pets  & Cars  & Caltech101 \\
 \cmidrule{2-6}          & Top 1 acc. & Top 1 acc. & Top 1 acc. & Top 1 acc. & Top 1 acc. \\
    \midrule
    CoOp~\cite{zhou2022learning}  & 66.10  & 30.97 & 82.77 & 60.20  & 90.26 \\
    CoCoOP~\cite{zhou2022conditional} & 67.23 & 31.72 & 83.14 & 59.78 & 90.43 \\
    Ours  & \textbf{71.13} & \textbf{47.34} & \textbf{90.60}  & \textbf{67.54} & \textbf{94.69} \\
    \bottomrule
    \end{tabular}}
  \label{tab:Comparison with CoOp}
\end{table}%

\begin{figure*}[t]
  \centering
\includegraphics[width=0.8\textwidth]{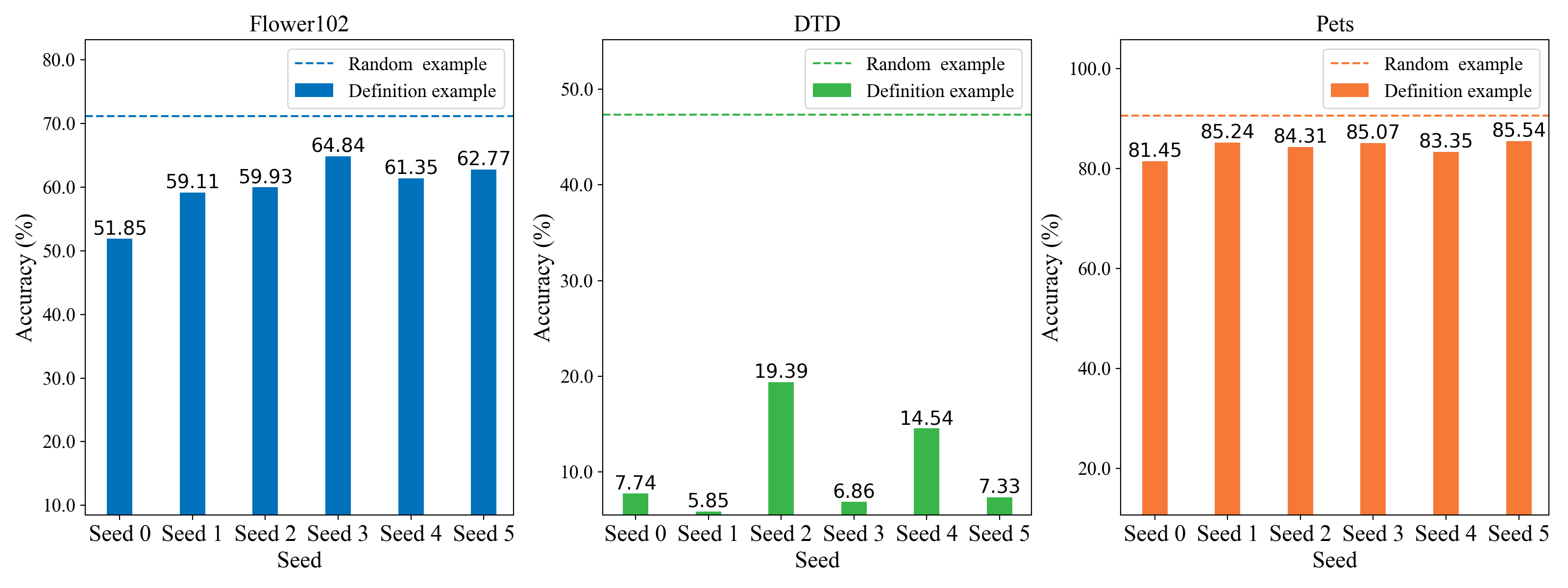}
  \caption{Experimental results w.r.t. different in-context example selection strategies.}
  \label{fig:random-definition}
\end{figure*}

\begin{table*}[t]
  \centering
  \caption{Comparative analysis of inference time and accuracy with existing TPT. The inference time (Infer. Time) is calculated in minutes.}
  \scalebox{0.6}{
    \begin{tabular}{lcccccrccccc}
    \toprule
    \multirow{2}[3]{*}{Method} & \multicolumn{2}{c}{Flower102} &       & \multicolumn{2}{c}{Pets} &       & \multicolumn{2}{c}{Cars} &       & \multicolumn{2}{c}{Caltech101} \\
\cmidrule{2-3}\cmidrule{5-6}\cmidrule{8-9}\cmidrule{11-12}          & Infer. Time $(\downarrow)$ & Top 1 acc. $(\uparrow)$  &       & Infer. Time $(\downarrow)$ & Top 1 acc. $(\uparrow)$   &       & Infer. Time $(\downarrow)$ &Top 1 acc. $(\uparrow)$   &       & Infer. Time $(\downarrow)$ & Top 1 acc. $(\uparrow)$ \\
\toprule
    TPT~\cite{shu2022test}   & 97.22 & 68.98 &       & 111.31 & 87.79 &       & 322.59 & 66.87 &       & 86.25 & 94.16 \\
    InCPL (Ours)  & \textbf{23.79} &\textbf{ 72.27} &       & \textbf{16.31} & \textbf{90.62} &       & \textbf{69.05} & \textbf{67.54} &       & \textbf{35.54} & \textbf{94.69} \\
    \bottomrule
    \end{tabular}}%
  \label{tab:inference time}
\end{table*}

\subsection{Additional Ablation Studies} \label{in-context-appendix}

\textbf{Comparison with CoOP and CoCoOP using the same examples.} We provide CoOp/CoCoOp’s results using the same examples as InCPL in Table~\ref{tab:Comparison with CoOp}. Experimental results show that our InCPL achieves better performance than CoOp and CoCoOp on this setting. 

\textbf{Training Strategy: In-context Learning \textit{vs.} Few-shot Learning.}
A conventional approach for leveraging the in-context examples dataset involves fine-tuning the prompt on a labeled dataset. As an alternative method, we implemented few-shot prompt learning using all available in-context examples and conducted a comparison with our InCPL approach. As demonstrated in Figure~\ref{fig:tsne}, the image representations learned through the few-shot approach exhibit lower resolution in distinguishing between classes compared to our prompted representations. The few-shot approach might introduce more variance and noise into the feature space, and the learned features are specific to the task rather than the individual test sample. In contrast, our InCPL method is specifically designed to learn an informative prompt for each test sample using in-context examples, enabling a more effective alignment between the sample and its associated examples.

\textbf{In-context Example Selection: Random \textit{vs.} Definition.} This experiment aims to evaluate two distinct approaches for selecting context examples to prompt each test sample: random and definition-based approaches. In the former approach, input-label pairs are randomly selected from candidate examples as in-context examples for each test sample, while the latter approach utilizes a common set of examples shared across all test samples once in-context examples have been sampled. As depicted in Figure ~\ref{fig:random-definition}, the definition-based selection approach also exhibits significant fluctuations with different random seeds, and its performance consistently lags behind the random-based selection approach. The primary reason for this discrepancy is that definition-based examples cannot consistently provide useful context information for all test samples. In contrast, random-based examples consistently offer each test sample domain-specific information relevant to the target distribution.

\textbf{Comparative Analysis of Inference Time Metrics with Existing TPT Method.} Table~\ref{tab:inference time} reports the comparative analysis of inference time metrics with TPT~\cite{shu2022test}. All experiments are conducted on one 2080 Ti GPU, and inference time (Infer. Time) is calculated in minutes. The table shows that TPT~\cite{shu2022test} needs significant inference time due to augmenting images by 64 times. In contrast, our InCPL only uses few in-context examples (i.e., 5) without any augmentation, requiring less inference time.

\section{Concusion}
In this paper, we tackle the challenge of model generalization in scenarios where the model lacks access to source training data but possesses only a very limited number of samples (e.g., just one sample for each category). Unlike few-shot learning, which utilizes only a small number of samples as training data, we treat these samples as domain-specific contextual information for each test sample. As a result, we introduce Visual In-Context Prompt Learning, a method that empowers a pre-trained vision-language model to make full use of in-context examples. Additionally, we have developed an effective language-aware prompt along with an example selection strategy and implemented a cyclic learning technique to facilitate the seamless integration of the vision prompt with the text prompt. Our experimental results across various downstream tasks consistently demonstrate the effectiveness of our approach.

\textbf{Limitation.} We develop a novel in-context evaluation paradigm that improves the performance of existing prompt learning algorithms in a ChatGPT-like manner. Under this paradigm, the in-context evaluator operates on test samples at test-time with unaltered model parameters, where the model only processes unlabeled test sample and in-context examples served as context information, without any other samples from the same category. 
The proposed InCPL requires a task-specific context dataset for each individual task. This dataset creation process inherently introduces certain priors, which can potentially limit the broad applicability of our proposed methods. In future work, we will consider developing unique context datasets tailored to different tasks. This approach would enable the adaptive selection of image-label pairs to serve as in-context prompt information for various tasks.
%
%
\bibliographystyle{splncs04}
\bibliography{reference}

\end{document}